\documentclass{article} % For LaTeX2e
\usepackage{iclr2019_conference,times}

\usepackage[utf8]{inputenc} % allow utf-8 input
\usepackage[T1]{fontenc}    % use 8-bit T1 fonts
\usepackage[pagebackref=true,breaklinks=true,colorlinks,citecolor=black,urlcolor=blue,bookmarks=false]{hyperref}    % hyperlinks
\usepackage{url}            % simple URL typesetting
\usepackage{booktabs}       % professional-quality tables
\usepackage{amsfonts}       % blackboard math symbols
\usepackage{nicefrac}       % compact symbols for 1/2, etc.
\usepackage{microtype}      % microtypography

\usepackage{enumitem}
\usepackage{footnote}
\makesavenoteenv{tabular}
\usepackage{fp}
\usepackage{xr}
\usepackage{color,xcolor}
\usepackage{epsfig}
\usepackage{graphicx}

\usepackage{adjustbox}
\usepackage{array}
\usepackage{booktabs}
\usepackage{colortbl}
\usepackage{float,wrapfig}
\usepackage{hhline}
\usepackage{array,multirow}
\usepackage{subcaption} % issues a warning with CVPR/ICCV format

\usepackage{amsmath,amsfonts,amsthm,amssymb}

\usepackage{bm}
\usepackage{nicefrac}
\usepackage{microtype}
\usepackage{dsfont}

\usepackage{changepage}
\usepackage{extramarks}
\usepackage{fancyhdr}
\usepackage{lastpage}
\usepackage{setspace}
\usepackage{soul}
\usepackage{xspace}
\usepackage{makecell}
\usepackage{mfirstuc}
\usepackage{url}

\usepackage{algorithm}
\usepackage{algpseudocode}
\usepackage{capt-of}
\usepackage{enumerate}
\usepackage{todonotes} % conflict with CVPR/ICCV format

\usepackage{titlesec}
\usepackage{caption}
\newcolumntype{L}[1]{>{\raggedright\let\newline\\\arraybackslash\hspace{0pt}}m{#1}}
\newcolumntype{C}[1]{>{\centering\let\newline\\\arraybackslash\hspace{0pt}}m{#1}}
\newcolumntype{R}[1]{>{\raggedleft\let\newline\\\arraybackslash\hspace{0pt}}m{#1}}

\newcommand{\hide}[1]{}

\newcommand{\reffig}[1]{Figure~\ref{fig:#1}}
\newcommand{\refsec}[1]{Section~\ref{sec:#1}}

\newcommand{\reftbl}[1]{Table~\ref{tbl:#1}}
\newcommand{\refalg}[1]{Algorithm~\ref{alg:#1}}

\newcommand{\refeq}[1]{Equation~\ref{eq:#1}}

\newcommand{\lblfig}[1]{\label{fig:#1}}
\newcommand{\lblsec}[1]{\label{sec:#1}}

\newcommand{\lbltbl}[1]{\label{tbl:#1}}

\newcommand{\ignorethis}[1]{}

\newcommand{\round}[2]{%
        \FPset{\a}{#1}
        \FPround{\a}{\a}{#2}
        \a
}
\newcommand{\musigprec}{2}
\newcommand{\psymbol}{\%}

\newcommand{\musigp}[2]{\round{#1}{\musigprec}\psymbol \pm\round{#2}{\musigprec}\psymbol}
\newcommand{\musigpb}[2]{\mathbf{\round{#1}{\musigprec}\psymbol} \pm\mathbf{\round{#2}{\musigprec}\psymbol}}
\newcommand{\p}[1]{\round{#1}{\musigprec}\psymbol}
\newcommand{\pb}[1]{\mathbf{\round{#1}{\musigprec}\psymbol}}

\makeatletter
\newcommand\fs@spaceruled{\def\@fs@cfont{\bfseries}\let\@fs@capt\floatc@ruled
  \def\@fs@pre{\vspace{-2\baselineskip}\hrule height.8pt depth0pt \kern2pt}%
  \def\@fs@post{\kern2pt\hrule\relax}%
  \def\@fs@mid{\kern2pt\hrule\kern2pt}%
  \let\@fs@iftopcapt\iftrue}
\makeatother

\algrenewcommand{\algorithmicrequire}{\textbf{Input:}}
\algrenewcommand{\algorithmicensure}{\textbf{Output:}}
\let\originalleft\left
\let\originalright\right
\renewcommand{\left}{\mathopen{}\mathclose\bgroup\originalleft}
\renewcommand{\right}{\aftergroup\egroup\originalright}

\newcommand{\V}[1]{\mathbf{#1}}

\newcommand{\loss}[1]{\ell(#1)}
\newcommand{\E}{\mathds{E}}
\newcommand{\dData}{\tilde{\V{x}}}
\newcommand{\dLr}{\tilde{\eta}}

\newcommand{\pascalvoc}[1]{PASCAL-VOC\xspace}
\newcommand{\pascalvocshort}[1]{PASCAL-VOC\xspace}
\newcommand{\imagenet}{ImageNet\xspace}
\newcommand{\birds}{CUB-200\xspace}
\newcommand{\birdsshort}{CUB-200\xspace}
\newcommand{\mnist}{MNIST\xspace}
\newcommand{\svhn}{SVHN\xspace}
\newcommand{\usps}{USPS\xspace}
\newcommand{\cifar}[1]{CIFAR#1\xspace}

\newcommand{\OURS}{Dataset Distillation\xspace}
\newcommand{\ours}{dataset distillation\xspace}
\newcommand{\image}{distilled image\xspace}
\newcommand{\data}{distilled data\xspace}
\newcommand{\images}{distilled images\xspace}

\newcommand{\Data}{Distilled data\xspace}
\newcommand{\Images}{Distilled images\xspace}

\newcommand{\ignore}[1]{}

\newcommand{\fixed}{fixed\xspace}
\newcommand{\stoch}{random\xspace}
\newcommand{\Fixed}{Fixed\xspace}
\newcommand{\Stoch}{Random\xspace}
\newcommand{\backp}{backpropagate\xspace}

\newcommand{\xavier}{Xavier~Initialization\xspace}
\newcommand{\kaiming}{He~Initialization\xspace}

\newcommand{\alexnet}{\textsc{AlexNet}\xspace}
\newcommand{\lenet}{\textsc{LeNet}\xspace}

\DeclareMathOperator*{\argmin}{arg\,min}

\makeatletter
\DeclareRobustCommand\onedot{\futurelet\@let@token\@onedot}
\def\@onedot{\ifx\@let@token.\else.\null\fi\xspace}

\def\eg{e.g\onedot}

\def\ie{i.e\onedot}

\def\wrt{w.r.t\onedot} 

\makeatother

\definecolor{MyDarkBlue}{rgb}{0,0.08,1}
\definecolor{MyDarkGreen}{rgb}{0.02,0.6,0.02}
\definecolor{MyDarkRed}{rgb}{0.8,0.02,0.02}
\definecolor{MyDarkOrange}{rgb}{0.40,0.2,0.02}
\definecolor{MyPurple}{RGB}{111,0,255}
\definecolor{MyRed}{rgb}{1.0,0.0,0.0}
\definecolor{MyGold}{rgb}{0.75,0.6,0.12}
\definecolor{MyDarkgray}{rgb}{0.66, 0.66, 0.66}

\setlist[itemize]{leftmargin=8mm}
\newcommand{\myparagraph}[1]{\vspace{-2pt}\noindent{\bf #1}}

\title{Dataset Distillation}

\author{%
\makebox[0.5\columnwidth][l]{Tongzhou Wang} \\
Facebook AI Research, MIT CSAIL \\
\And%
\makebox[0.5\columnwidth][l]{Jun-Yan Zhu} \\
MIT CSAIL \\
\AND%
\makebox[0.5\columnwidth][l]{Antonio Torralba} \\
MIT CSAIL \\
\And%
\makebox[0.5\columnwidth][l]{Alexei A. Efros} \\
UC Berkeley \\
}

\iclrfinalcopy % Uncomment for camera-ready version, but NOT for submission.
\begin{document}

\maketitle

%auto-ignore
%!TEX root = ../main.tex

\begin{abstract}
Model distillation aims to distill the knowledge of a complex model into a simpler one.  In this paper, we consider an alternative formulation called {\em dataset distillation}: we keep the model fixed and instead attempt to distill the knowledge from a large training dataset into a small one.  The idea is to {\em synthesize} a small number of data points that do not need to come from the correct data distribution, but will, when given to the learning algorithm as training data, approximate the model trained on the original data.  For example, we show that it is possible to compress $60,000$ \mnist training images into just $10$ synthetic {\em distilled images} (one per class) and achieve close to original performance with only a few gradient descent steps, given a \fixed network initialization. We evaluate our method in various initialization settings and with different learning objectives. Experiments on multiple datasets show the advantage of our approach compared to alternative methods.
\end{abstract}
%auto-ignore
%!TEX root = ../main.tex

\section{Introduction}
\lblsec{intro}
\citet{hinton2015distilling} proposed network distillation as a way to transfer the knowledge from an ensemble of many separately-trained networks into a single, typically compact network, performing a type of model compression.
In this paper, we are considering a related but orthogonal task: rather than distilling the model, we propose to distill the dataset. Unlike network distillation, we keep the model fixed but encapsulate the knowledge of the entire training dataset, which typically contains thousands to millions of images, into a small number of  synthetic training images. We show that we can go as low as {\em one} synthetic image per category, training the same model to reach surprisingly good performance on these synthetic images. For example in~\reffig{teaser}a, we compress  $60,000$ training images of \mnist digit dataset into only $10$ synthetic images (one per class), given a fixed network initialization.  Training the standard \lenet~\citep{lecun1998gradient} on these $10$ images yields test-time \mnist recognition performance of $94\%$, compared to $99\%$ for the original dataset. For networks with unknown \stoch weights, $100$ synthetic images train to $80\%$ with a few gradient descent steps. We name our method \emph{\OURS} and these images \emph{\images}.

But why is dataset distillation useful?  There is the purely scientific question of how much data is encoded in a given training set and how compressible it is?
Moreover, given a few \images, we can now ``load up" a given network with an entire dataset-worth of knowledge much more efficiently, compared to traditional training that often uses tens of thousands of gradient descent steps.

A key question is whether it is even possible to compress a dataset into a small set of synthetic data samples. For example, is it possible to train an image classification model on synthetic images out of the manifold of natural images? Conventional wisdom would suggest that the answer is no, as the synthetic training data may not follow the same distribution of the real test data.  Yet, in this work, we show that this is indeed possible.
We present a new optimization algorithm for synthesizing a small number of synthetic data samples not only capturing much of the original training data but also tailored explicitly for fast model training in only a few gradient steps.
To achieve our goal, we first derive the network weights as a differentiable function of our synthetic training data. Given this connection, instead of optimizing the network weights for a particular training objective, we  optimize the pixel values of our \images. However, this formulation requires access to the initial weights of the network. To relax this assumption, we develop a method for generating \images for randomly initialized networks. To further boost performance, we propose an iterative version, where we obtain a sequence of \images and these \images can be trained with multiple epochs (passes). Finally, we study a simple linear model, deriving a lower bound on the size of distilled data required to achieve the same performance as training on the full dataset.

%auto-ignore
%!TEX root = ../main.tex

\begin{figure}[t]
    \centering
    \vspace{-20pt}
    \includegraphics[width=0.985\textwidth]{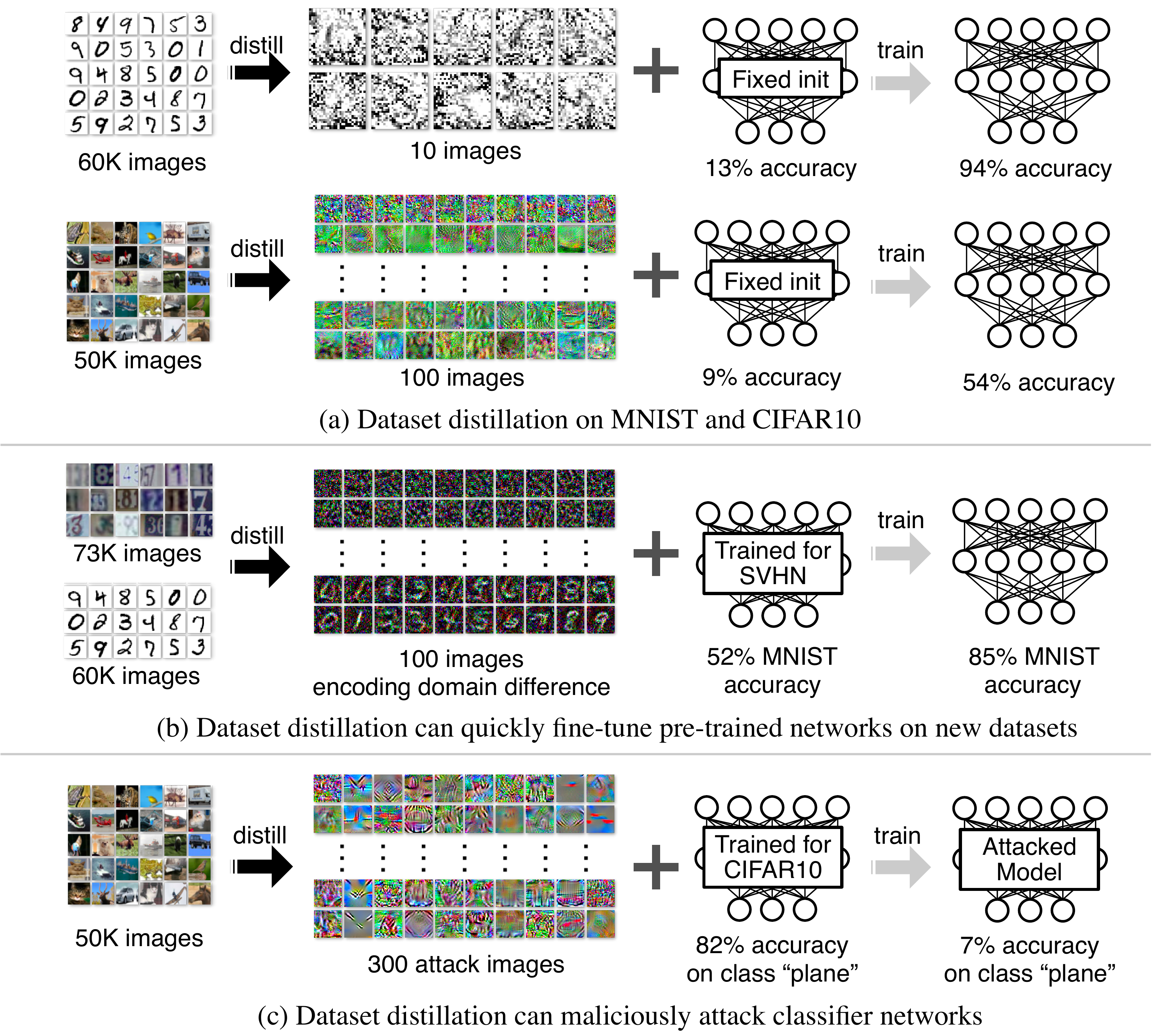}\vspace{-3pt}
    \caption{We distill the knowledge of tens of thousands of images into a few synthetic training images called  \images.
    (a) On \mnist, $10$ \images can train a standard \lenet with a fixed initialization to $94\%$ test accuracy,  compared to $99\%$ when fully trained. On \cifar{10}, $100$ \images can train a  network with a fixed initialization to $54\%$ test accuracy, compared to $80\%$ when fully trained.
    (b) We distill the domain difference between \svhn and \mnist into $100$ images. These images can quickly fine-tune pre-trained \svhn networks to achieve high accuracy on \mnist.
    (c) Our formulation can create dataset poisoning images. After trained with these images for {\em one single} gradient step,  networks  will catastrophically misclassify one category.}
    \lblfig{teaser}
\end{figure}

We demonstrate that a handful of \images can be used to train a model with a \fixed initialization to achieve surprisingly high performance. For networks  pre-trained on other tasks, our method can  find \images for fast model fine-tuning.  We test our method on several initialization settings: fixed initialization, \stoch initialization, fixed pre-trained weights, and \stoch pre-trained weights, as well as two training objectives: image classification and malicious dataset poisoning attack. Extensive experiments on four publicly available datasets, MNIST, CIFAR10, PASCAL-VOC, and CUB-200, show that our approach often outperforms  existing methods. Please check out our \href{https://github.com/SsnL/dataset-distillation}{code} and \href{https://ssnl.github.io/dataset_distillation}{website} for more details.

%auto-ignore
%!TEX root = ../main.tex

\section{Related Work}
\lblsec{related}
\myparagraph{Knowledge distillation.}
The main inspiration for this paper is network distillation~\citep{hinton2015distilling}, a widely used technique in ensemble learning~\citep{radosavovic2017data} and model compression~\citep{ba2014deep,romero2014fitnets,howard2017mobilenets}. While network distillation aims to distill the knowledge of multiple networks into a single model, our goal is to compress the knowledge of an entire dataset into a few synthetic training images. Similar to our approach, data-free knowledge distillation also optimizes synthetic data samples, but with a different objective of matching activation statistics of a teacher model in knowledge distillation~\citep{lopes2017data}. Our method is also related to the theoretical concept of teaching dimension, which specifies the size of dataset necessary to teach a target model to a learner \citep{shinohara1991teachability,goldman1995complexity}. However, methods~\citep{zhu2013machine,zhu2015machine} inspired by this concept need the existence of target models, which our method does not require.

\myparagraph{Dataset pruning, core-set construction, and instance selection.} Another way to distill knowledge is to summarize the entire dataset by a small subset, either by only using the ``valuable''  data for model training~\citep{angelova2005pruning,felzenszwalb2010object, lapedriza2013all} or by only labeling the ``valuable'' data via active learning~\citep{cohn1996active,tong2001support}. Similarly, core-set construction \citep{tsang2005core,har2007smaller,bachem2017practical,sener2018active} and instance selection \citep{olvera2010review} methods aim to select a subset of the entire training data, such that models trained on the subset will perform as well as the model trained on full dataset. For example, solutions to many classical linear learning algorithms, \eg, Perceptron~\citep{rosenblatt1957perceptron} and SVMs~\citep{hearst1998support}, are weighted sums of a subset of training examples, which can be viewed as core-sets. However, algorithms constructing these subsets require many more training examples per category than we do, in part because their ``valuable'' images have to be real, whereas our \images are exempt from this constraint.

\myparagraph{Gradient-based hyperparameter optimization.}
Our work bears similarity with gradient-based hyperparameter optimization techniques, which compute the gradient of hyperparameter \wrt the final validation loss by reversing the entire training procedure~\citep{bengio2000gradient,domke2012generic,maclaurin2015gradient,pedregosa2016hyperparameter}. We also \backp errors through optimization steps. However, we use only training set data and focus more heavily on learning synthetic training data rather than tuning hyperparameters. To our knowledge, this direction has only been slightly touched on previously~\citep{maclaurin2015gradient}. We explore it in greater depth and demonstrate the idea of \ours in various settings. More crucially, our \images work well across \stoch initialization weights, not possible by prior work.

\myparagraph{Understanding datasets.}
Researchers have presented various approaches for understanding and visualizing learned models~\citep{zeiler2014visualizing,zhou2014object,mahendran15understanding,bau2017network, Koh2017understanding}. Unlike these approaches, we are interested in understanding the intrinsic properties of the training data rather than a specific trained model.
Analyzing training datasets has, in the past, been mainly focused on the investigation of bias in datasets~\citep{ponce2006dataset,torralba2011unbiased}.  For example, \citet{torralba2011unbiased} proposed to quantify the ``value'' of dataset samples using cross-dataset generalization.
Our method offers a new perspective for understanding datasets by distilling full datasets into a few synthetic samples.

%auto-ignore
%!TEX root = ../main.tex
\section{Approach}
\lblsec{method}

Given a model and a dataset, we aim to obtain a new, much-reduced \emph{synthetic} dataset which performs almost as well as the original dataset. We first present our main optimization algorithm for training a network with a \fixed initialization with one gradient descent (GD) step (\refsec{opt}).
In \refsec{random}, we derive the resolution to a more challenging case, where initial weights are \stoch rather than \fixed. In \refsec{linear}, we further study a linear network case to help readers understand both the property and limitation of our method. We also discuss the initial weights distribution with which our method can work well. In \refsec{multi}, we extend our approach to more than one gradient descent steps and more than one epoch (pass). Finally, \refsec{diff-init} and \refsec{diff-obj} demonstrate how to obtain \images with different initialization distributions and learning objectives.

Consider a training dataset $\V{x} = \{x_i\}_{i=1}^N$, we parameterize our neural network as $\theta$ and denote  $\loss{x_i, \theta}$ as the loss function that represents the loss of this network on a data point $x_i$. Our task is to find the minimizer of the empirical error over entire training data:
\begin{equation}
    \theta^* = \argmin_{\theta} \frac{1}{N} \sum_{i=1}^{N} \loss{x_i, \theta} \triangleq  \argmin_{\theta} \loss{\V{x}, \theta},
\end{equation}
where for notation simplicity we overload the $\loss{\cdot}$ notation so that $\loss{\V{x}, \theta}$ represents the average error of $\theta$ over the entire dataset. We make the mild assumption that $\ell$ is twice-differentiable, which holds true for the majority of modern machine learning models and tasks.

\subsection{Optimizing Distilled Data}
\lblsec{opt}
Standard training usually applies minibatch stochastic gradient descent or its variants. At each step $t$, a minibatch of training data $\V{x}_t = \{x_{t,j}\}_{j=1}^n$ is sampled to update the current parameters as
\begin{align*}
    \theta_{t + 1} & = \theta_t - \eta \nabla_{\theta_t} \loss{\V{x}_t, \theta_t},
\end{align*} where $\eta$ is the learning rate. Such a training process often takes tens of thousands or even millions of update steps to converge. Instead, we aim to learn a tiny set of synthetic distilled training data $\dData = \{\tilde{x}_i \}_{i=1}^M$
with $M \ll N$ and a corresponding learning rate $\dLr$ so that a single GD step such as
\begin{equation}
    \theta_1 = \theta_0 - \dLr \nabla_{\theta_0} \loss{\dData, \theta_0} \label{eq:sc_next_theta}
\end{equation} using these learned synthetic data $\dData$ can greatly boost the performance on the real test set. Given an initial $\theta_0$, we obtain these synthetic data $\dData$ and learning rate $\dLr$ by minimizing the objective below $\mathcal{L}$:
\begin{align}
    \dData^*, \dLr^* = \argmin_{\dData, \dLr} \mathcal{L}(\dData, \dLr; \theta_0)
     = \argmin_{\dData, \dLr} \loss{\V{x}, \theta_1}
     = \argmin_{\dData, \dLr} \loss{\V{x}, \theta_0 - \dLr \nabla_{\theta_0} \loss{\dData, \theta_0}}, \label{eq:sc_objective_expand}
\end{align}
where we derive the new weights $\theta_1$ as a function of \data $\dData$ and learning rate $\dLr$ using~\refeq{sc_next_theta} and then evaluate the new weights over all the training data $\V{x}$.
The loss $\mathcal{L}(\dData, \dLr; \theta_0)$ is differentiable \wrt $\dData$ and $\dLr$, and can thus be optimized using standard gradient-based methods. In many classification tasks, the data $\V{x}$ may contain discrete parts, \eg, class labels in data-label pairs. For such cases, we fix the discrete parts rather than learn them.

\subsection{Distillation for Random Initializations}
\lblsec{random}

Unfortunately, the above \data optimized for a given initialization do not generalize well to other initializations. The \data often look like random noise (\eg, in~\reffig{charge_one_MNIST}) as it encodes the information of both training dataset $\V{x}$ and a particular network initialization $\theta_0$. To address this issue, we turn to calculate a small number of \data that can work for networks with \stoch initializations from a specific distribution. We formulate the optimization problem as follows:
\begin{equation}
  \dData^*, \dLr^* = \argmin_{\dData, \dLr} \E_{\theta_0 \sim p(\theta_0)} \mathcal{L}(\dData, \dLr; \theta_0), \label{eq:random}
\end{equation}
where the network initialization $\theta_0$ is randomly sampled from a distribution $p(\theta_0)$.
During our optimization, the \data are optimized to work well for randomly initialized networks. \refalg{single} illustrates our main method. In practice, we observe that the final \data generalize well to  unseen initializations. In addition, these \images often look quite informative, encoding the discriminative features of each category (\eg, in~\reffig{charge_random_10}).

For \data to be properly learned, it turns out crucial for $\loss{\V{x}, \cdot}$ to share similar local conditions (e.g., output values, gradient magnitudes) over initializations $\theta_0$ sampled from $p(\theta_0)$. In the next section, we derive a lower bound on the size of \data needed for a simple model with arbitrary initial $\theta_0$, and discuss its implications on choosing $p(\theta_0)$.

%auto-ignore
%!TEX root = ../main.tex

\begin{algorithm}[t]
    \small
    \caption{\OURS}
    \label{alg:single}
    \begin{algorithmic}[1]
        \Require {$p(\theta_0)$: distribution of initial weights; $M$: the number of \data}
        \Require {$\alpha$: step size; $n$: batch size; $T$: the number of optimization iterations; $\dLr_0$: initial value for $\dLr$}
        \State {Initialize $\dData = \{\tilde{x}_i\}_{i=1}^M$ randomly, $\dLr \gets \dLr_0$}
        \For {\textbf{each} training step $t = 1$ to $T$ }
            \State {Get a minibatch of real training data $\V{x}_t = \{x_{t, j}\}_{j=1}^n$}
            \State {Sample a batch of initial weights $\theta^{(j)}_0 \sim p(\theta_0)$}
            \For {\textbf{each} sampled $\theta^{(j)}_0$}
                \State Compute updated parameter with GD: $\theta_1^{(j)} = \theta_0^{(j)} - \dLr \nabla_{\theta_0^{(j)}} \loss{\dData, \theta_0^{(j)}}$ \label{alg:line:gd_step}
                \State {Evaluate the objective function on real training data: $\mathcal{L}^{(j)} = \loss{\V{x}_t, \theta_1^{(j)}}$} \label{alg:line:objective_fn}
            \EndFor
            \State {Update $\dData \gets \dData - \alpha \nabla_{\dData} \sum_j \mathcal{L}^{(j)}$, and $\dLr \gets \dLr - \alpha \nabla_{\dLr} \sum_j \mathcal{L}^{(j)}$} \label{alg:line:bwd_through_opt}
        \EndFor
        \Ensure {\data $\dData$ and optimized learning rate $\dLr$}
    \end{algorithmic}
\end{algorithm}%

\subsection{Analysis of a Simple Linear Case}
\lblsec{linear}
This section studies our formulation in a simple linear regression problem with quadratic loss. We derive the lower bound of the size of \data needed to achieve the same performance as training on the full dataset for arbitrary initialization with one GD step.
Consider a dataset $\V{x}$ containing $N$ data-target pairs $\{(d_i, t_i)\}_{i=1}^{N}$, where  $d_i\in \mathbb{R}^{D}$ and $t_i\in \mathbb{R}$, which we represent as two matrices: an $N\times D$ data matrix $\V{d}$ and an $N\times 1$ target matrix $\V{t}$. Given  mean squared error and a $D\times 1$ weight matrix $\theta$, we have
\begin{equation}
    \loss{\V{x}, \theta} = \loss{(\V{d}, \V{t}), \theta} = \frac{1}{2N} \lVert \V{d} \theta - \V{t} \rVert^2.
\end{equation}
We aim to learn $M$ synthetic data-target pairs $\dData = (\tilde{\V{d}}, \tilde{{\V{t}}})$, where $\tilde{\V{d}}$ is an $M\times D$ matrix, $\tilde{\V{t}}$ an $M\times 1$ matrix ($M \ll N$), and $\dLr$ the learning rate, to minimize $\loss{\V{x}, \theta_0 - \dLr \nabla_{\theta_0} \loss{\dData, \theta_0}}$.  The updated weight matrix after one GD step with these distilled data is \begin{equation}
    \theta_1 =\theta_0 - \dLr \nabla_{\theta_0} \loss{\dData, \theta_0}
     = \theta_0 - \frac{\dLr}{M} \tilde{\V{d}}^T (\tilde{\V{d}} \theta_0 - \tilde{\V{t}})
     = (\V{I} - \frac{\dLr}{M} \tilde{\V{d}}^T \tilde{\V{d}}) \theta_0 + \frac{\dLr}{M} \tilde{\V{d}}^T \tilde{\V{t}}.  \label{eq:linear-theta1}
\end{equation}
For the quadratic loss, there always exists learned distilled data $\dData$ that can achieve the same performance as training on the full dataset $\V{x}$ (i.e., attaining the global minimum) for {\em any} initialization $\theta_0$. For example, given any global minimum solution $\theta^*$, we can choose $\tilde{\V{d}} = N \cdot \V{I}$ and $\tilde{\V{t}} = N \cdot \theta^*$. But how small can the size of \data be? For such models, the global minimum is attained at any $\theta^*$ satisfying $\V{d}^T \V{d} \theta^* = \V{d}^T \V{t}$. Substituting Equation~\eqref{eq:linear-theta1} in the condition above, we have
\begin{equation}
    \V{d}^T \V{d} (\V{I} - \frac{\dLr}{M} \tilde{\V{d}}^T \tilde{\V{d}}) \theta_0 + \frac{\dLr}{M} \V{d}^T \V{d} \tilde{\V{d}}^T \tilde{\V{t}} = \V{d}^T \V{t}.
\end{equation}
Here we make the mild assumption that the feature columns of the data matrix $\V{d}$ are independent (\ie, $\V{d}^T \V{d}$ has full rank).
For a $\dData = (\tilde{\V{d}}, \tilde{\V{t}})$ to satisfy the above equation for any $\theta_0$, we must have
\begin{equation}
 \V{I} - \frac{\dLr}{M} \tilde{\V{d}}^T \tilde{\V{d}} = \V{0},
\end{equation}which implies that $\tilde{\V{d}}^T \tilde{\V{d}}$ has full rank and $M \geq D$.

\myparagraph{Discussion.} The analysis above only considers a simple case but suggests that any small number of \data fail to generalize to arbitrary initial $\theta_0$. This is intuitively expected as the optimization target $\loss{\V{x}, \theta_1} = \loss{\V{x}, \theta_0 - \dLr \nabla_{\theta_0} \loss{\dData, \theta_0}}$ depends on the local behavior of $\loss{\V{x}, \cdot}$ around $\theta_0$, which can be drastically different across various initializations $\theta_0$. The lower bound $M \geq D$ is a quite restricting one, considering that real datasets often have thousands to even hundreds of thousands of dimensions (\eg, images). This analysis motivates us to focus on $p(\theta_0)$ distributions that yield similar local conditions.  \refsec{diff-init} explores several practical choices. To address the limitation of using a single GD step,  we extend our method to multiple GD steps in the next section.

\subsection{Multiple Gradient Descent Steps and Multiple Epochs}
\lblsec{multi}
We extend \refalg{single} to more than one gradient descent steps by changing Line~\ref{alg:line:gd_step} to multiple sequential GD steps each on a different batch of \data and learning rates, \ie, each step $i$ is
\begin{equation}
    \theta_{i + 1} = \theta_i - \dLr_i \nabla_{\theta_i} \loss{\dData_i, \theta_i},
\end{equation}
and changing Line~\ref{alg:line:bwd_through_opt} to \backp through all steps. However, naively computing gradients is memory and computationally expensive. Therefore, we exploit a recent technique called back-gradient optimization, which allows for significantly faster gradient calculation of such updates in reverse-mode differentiation. Specifically, back-gradient optimization formulates the necessary second order terms into efficient Hessian-vector products~\citep{pearlmutter1994fast}, which can be easily calculated with modern automatic differentiation systems such as PyTorch~\citep{paszke2017automatic}. For further  algorithmic details, we refer readers to prior work~\citep{domke2012generic,maclaurin2015gradient}.

\myparagraph{Multiple epochs.} To further improve the performance, we train the network for multiple epochs (passes) over the same sequence of \data.
In other words, for each epoch, our method cycles through all GD steps, where each step is associated with a batch of \data. We do not tie the trained learning rates across epochs as later epochs often use smaller learning rates. In \refsec{expr:init}, we find that using multiple steps and multiple epochs is more effective than using just one on neural networks, with the total amount of \data fixed.

\subsection{Distillation with Different Initializations}
\lblsec{diff-init}

Inspired by the analysis of the simple linear case in \refsec{linear}, we aim to focus on initial weights distributions $p(\theta)$ that yield similar local conditions over the data distribution. In this work, we focus on the following four practical choices:
\begin{itemize}
    \item \myparagraph{\Stoch initialization:} Distribution over random initial weights, \eg, \kaiming~\citep{he2015delving} and \xavier~\citep{glorot2010understanding} for neural networks.
    \item \myparagraph{\Fixed initialization:} A particular \fixed network initialized by the method above.
    \item \myparagraph{\Stoch pre-trained weights:} Distribution over models pre-trained on other tasks or datasets, \eg, \alexnet~\citep{krizhevsky2012imagenet} networks trained on \imagenet \citep{deng2009imagenet}.
    \item \myparagraph{\Fixed pre-trained weights:} A particular \fixed network pre-trained on other tasks and datasets.
\end{itemize}

\myparagraph{Distillation with pre-trained weights.} Such learned \data essentially fine-tune weights pre-trained on one dataset to perform well for a new dataset, thus bridging the gap between the two domains. Domain mismatch and dataset bias represent a challenging problem in machine learning today~\citep{torralba2011unbiased}.
Extensive prior work has been proposed to adapt models to new tasks and datasets~\citep{daume2007frustratingly,saenko2010adapting}.
In this work, we characterize the domain mismatch via \data. In \refsec{app_results}, we show that a small number of \images are sufficient to quickly adapt CNN models to new datasets and tasks.

\subsection{Distillation with Different Objectives}
\lblsec{diff-obj}
%Previous sections show that we can train \data to minimize the loss of the distilled task $\loss{\V{x}, \theta_1}$ defined on the final updated weights $\theta_1$ (Line~\ref{alg:line:objective_fn} in Algorithm~\ref{alg:single}).
\Data learned with different learning objectives can train models to exhibit different desired behaviors. We have already mentioned image classification as one of the applications, where \images help to train accurate classifiers. Below, we introduce a different learning objective to further demonstrate the flexibility of our method.

\myparagraph{Distillation for malicious data poisoning.}
For example, our approach can be used to construct a new form of data poisoning attack. To illustrate this idea, we consider the following scenario. When a single GD step is applied with our synthetic adversarial data, a well-behaved image classifier catastrophically forgets one category but still maintains high accuracy on other categories.

Formally, given an attacked category $K$ and a target category $T$, %we want the classifier to misclassify images of category $K$ as category $T$. To achieve this,
we minimize a new objective $\ell_{K \rightarrow T}(\V{x}, \theta_1)$, which is a classification loss  that encourages $\theta_1$ to misclassify images of category $K$  as category $T$ while correctly predicting other images, \eg, a cross entropy loss with target labels of $K$ modified to $T$. Then, we can obtain the malicious \images by optimizing
\begin{equation}
    \dData^*, \dLr^* = \argmin_{\dData, \dLr} \E_{\theta_0 \sim p(\theta_0)} \mathcal{L}_{K \rightarrow T}(\dData, \dLr; \theta_0),
\end{equation}where $p(\theta_0)$ is the distribution over \textit{\stoch weights} of well-optimized classifiers. Trained on a distribution of such classifiers, the \images \textit{do not} require access to the exact model weights and thus can generalize to unseen models. In our experiments, the malicious \images are trained on $2000$ well-optimized models and evaluated on 200 held-out ones.

Compared to prior data poisoning attacks \citep{biggio2012poisoning,li2016data,munoz2017towards,Koh2017understanding}, our approach crucially {\it does not} require the poisoned training data to be stored and trained on repeatedly. Instead, our method attacks the model training in one iteration and with only a few data. This advantage makes our method potentially effective for online training algorithms and useful for the case where malicious users hijack the data feeding pipeline for only one gradient step (\eg, one network transmission). In \refsec{app_results}, we show that a single batch of \data applied in one step can successfully attack well-optimized neural network models. This setting can be viewed as distilling the knowledge of a specific category into data.

%auto-ignore
%!TEX root = ../main.tex

\section{Experiments}
\lblsec{experiments}
We report image classification results on \mnist~\citep{lecun1998mnist} and \cifar{10}~\citep{cifar10}. For \mnist, the \images are trained with \lenet~\citep{lecun1998gradient}, which achieves about $99\%$ test accuracy if fully trained. For \cifar{10}, we use a network architecture~\citep{krizhevsky2012cuda} that achieves around $80\%$ test accuracy if fully trained. For \stoch initializations and \stoch pre-trained weights, we report means and standard deviations over $200$ held-out models, unless otherwise specified. The \href{https://github.com/SsnL/dataset-distillation}{code} and full results can be found at our \href{https://github.com/SsnL/dataset-distillation}{website}.

\myparagraph{Baselines.} For each experiment, in addition to baselines specific to the setting, we generally compare our method against baselines trained with data derived or selected from real training images:
\begin{itemize}
\item \myparagraph{Random real images:} We randomly sample the same number of real images per category.
\item \myparagraph{Optimized real images:} We sample different sets of random real images as above, and choose the top $20\%$ best performing sets.
\item \myparagraph{$k$-means:} We apply $k$-means clustering to each category, and use the cluster centroids as training images.
\item \myparagraph{Average real images:} We compute the average image for each category, which is reused in different GD steps.
\end{itemize}
For these baselines, we perform each evaluation on $200$ held-out models with all combinations of $\text{learning rate} \in \{\text{learned learning rate with our method}, 0.001, 0.003, 0.01, 0.03, 0.1, 0.3\}$ and $\text{\#epochs} \in \{1, 3, 5\}$, and report results using the best performing combination. Please see the appendix \refsec{training_details} for more details about training and baselines.

\subsection{\OURS}
\lblsec{expr:init}
\begin{figure}
\centering
\input{figText/charge_fixed}
\input{figText/charge_random}
\end{figure}
\begin{figure}
\centering
\input{figText/ablation}
\input{figText/charge-one-vs-N-steps}
\end{figure}

\myparagraph{\Fixed initialization.} With access to  initial network weights, \images can directly train a  fixed network to reach high performance. For example, $10$ \images can boost the performance of a neural network with an initial accuracy $\p{12.9}$ to a final accuracy $\p{93.76}$ on \mnist (\reffig{charge_one_MNIST}).
Similarly, $100$ images can train a network with an initial accuracy $\p{8.82}$ to $\p{54.03}$ test accuracy on \cifar{10} (\reffig{charge_one_Cifar}).
This result suggests that even only a few \images have enough capacity to distill part of the dataset.

\myparagraph{\Stoch initialization.} Trained with randomly sampled initializations using \xavier~\citep{glorot2010understanding}, the learned \images do not need to encode information tailored for a particular starting point and thus can represent meaningful content independent of network initializations. In \reffig{charge_random_10}, we see that such \images reveal the discriminative features of corresponding categories: \eg, the {\tt ship} image in \reffig{charge_random_Cifar}. These $100$ images can train randomly initialized networks to $\p{36.79}$ average test accuracy on \cifar{10}. Similarly, for \mnist, the $100$ \images shown in \reffig{charge_random_MNIST} can train randomly initialized networks to $\p{79.50}$ test accuracy.

\myparagraph{Multiple gradient descent steps and multiple epochs.} In \reffig{charge_random_10}, \images are learned for $10$ GD steps applied in $3$ epochs, leading to a total of $100$ images (with each step containing one image per category). Images used in  early steps tend to look noisier. However, in later steps, the \images gradually look like real data and share the discriminative features for these categories. \reffig{ablation_steps} shows that using more steps significantly improves the results. \reffig{ablations_epochs} shows a similar but slower trend as the number of epochs increases.
We observe that longer training (\ie, more epochs) can help the model learn all the knowledge from the \images, but the performance is eventually limited by the total number of images. Alternatively, we can train the model with one GD step but a big batch size. \refsec{linear} has shown theoretical limitations of using only one step in a simple linear case. In \reffig{charge-1-vs-n}, we  observe that using multiple steps drastically outperforms the single step method, given the same number of \images.

\begin{table}
%auto-ignore
%!TEX root = ../main.tex

\vspace{-1pt}
\centering
\resizebox{\columnwidth}{!}{%
\renewcommand{\arraystretch}{1.2}
\renewcommand{\musigprec}{1}
\renewcommand{\psymbol}{}
% Baselines:
%   GD:
%     random train
%     opt random train
%     kmeans train
%     average train
%  k-NN:
%     random train
%     kmeans
\begin{tabular}{|c|c|c|c|c|c|c|c|c|}
    \hline
    & \multicolumn{2}{c|}{Ours} & \multicolumn{6}{c|}{Baselines} \\
    \cline{2-9}
    & \multirow{2}{*}{\Fixed init.} &  \multirow{2}{*}{\Stoch init.} & \multicolumn{4}{c|}{Used as training data in same number of GD steps} & \multicolumn{2}{c|}{Used in $\mathtt{K}$-NN classification} \\
    \cline{4-9}
    & & & Random real & Optimized real & $k$-means & Average real & Random real & $k$-means \\
    \hline

    \mnist & $\pb{96.62}$ & $\musigp{79.50}{8.08}$
    & $\musigp{68.55}{9.78}$ & $\musigp{73.01}{7.63}$ & $\musigp{76.43}{9.51}$
    & $\musigp{77.09}{2.70}$ & $\musigp{71.53}{2.06}$ & $\musigpb{92.19}{0.14}$ \\
    \hline
    \cifar{10}
    & $\pb{54.03}$ & $\musigpb{36.79}{1.18}$
    & $\musigp{21.30}{1.47}$ & $\musigp{23.40}{1.33}$ & $\musigp{22.48}{3.09}$
    & $\musigp{22.34}{0.65}$ & $\musigp{18.82}{1.28}$ & $\musigp{29.42}{0.28}$ \\
    \hline
\end{tabular}
}
% \vspace{-5pt}%
\caption{Comparison between our method trained for ten GD steps and three epochs and various baselines. For baselines using $\mathtt{K}$-Nearest Neighbor ($\mathtt{K}$-NN), best result among all combinations of $\text{distance metric} \in \{l_1, l_2\}$ and $\mathtt{K} \in \{1, 3\}$ is reported. In $\mathtt{K}$-NN and $k$-means, $\mathtt{K}$ and $k$ can have different values. All methods use ten images per category (100 in total), except for the average real images baseline, which reuses the same images in different GD steps.
}
\lbltbl{charge}
\vspace{-3pt}

\end{table}%
\reftbl{charge} summarizes the results of our method and all baselines. Our method with both \fixed and \stoch initializations outperforms all the baselines on \cifar{10} and most of the baselines on \mnist.

\subsection{Distillation with Different Initializations and Objectives}
\lblsec{app_results}

Next, we show two extended settings of our main algorithm discussed in \refsec{diff-init} and \refsec{diff-obj}. Both cases assume that the initial weights are \stoch but pre-trained on the same dataset. We train the \images on $2000$ \stoch pre-trained models and evaluate them on {\it unseen} models.

\myparagraph{\Fixed and \stoch pre-trained weights on digits.} As shown in \refsec{diff-init}, we can optimize \images to quickly fine-tune pre-trained models for a new dataset.
\reftbl{adapt} shows that our method is more effective than various baselines on adaptation between three digits datasets: \mnist, \usps \citep{hull1994database}, and \svhn \citep{netzer2011reading}. We also compare our method against a state-of-the-art few-shot domain adaptation method \citep{motiian2017few}. Although our method uses the entire training set to compute the \images, both methods use the same number of images to distill the knowledge of target dataset. Prior work~\citep{motiian2017few} is outperformed by our method with \fixed pre-trained weights on all the tasks, and by our method with \stoch pre-trained weights on two of the three tasks. This result shows that our \images effectively compress the information of target datasets.

\begin{table}[t]%
%auto-ignore
%!TEX root = ../main.tex

% \vspace{-3pt}
\centering
\resizebox{1.0\columnwidth}{!}{%
\renewcommand{\arraystretch}{1.2}
\renewcommand{\musigprec}{1}
\renewcommand{\psymbol}{}

\begin{tabular}{| c|c|c|c|c|c|c||c||c|c|}
    \hline
    &
    \multirow{3}{*}{\shortstack{Ours\\w/ \fixed\\pre-trained}} &
    \multirow{3}{*}{\shortstack{Ours\\w/ \stoch\\pre-trained}} &
    \multirow{3}{*}{Random real} &
    \multirow{3}{*}{Optimized real} &
    \multirow{3}{*}{$k$-means} &
    \multirow{3}{*}{Average real} &
    \multirow{3}{*}{\shortstack{Adaptation\\\citeauthor{motiian2017few}\\\citeyearpar{motiian2017few}}} &
    \multirow{3}{*}{No adaptation} &
    \multirow{3}{*}{\shortstack{Train on \textbf{full}\\target dataset}} \\
    & & & & & & & & & \\
    & & & & & & & & & \\
    \hline

    $\mathcal{M} \rightarrow \mathcal{U}$ &
    $\pb{97.90}$ & $\musigpb{95.38}{1.81}$ &
    $\musigp{94.89}{0.80}$ &
    $\musigp{95.16}{0.69}$ &
    $\musigp{92.18}{1.64}$ &
    $\musigp{93.89}{0.83}$ &
    $\musigpb{96.65}{0.45}$ &
    $\musigp{90.43}{2.97}$ & $\musigp{97.32}{0.27}$ \\
    \hline

    $\mathcal{U} \rightarrow \mathcal{M}$ &
    $\pb{93.19}$ & $\musigpb{92.74}{1.38}$ &
    $\musigp{87.05}{2.88}$ &
    $\musigp{87.59}{2.14}$ &
    $\musigp{85.62}{3.13}$ &
    $\musigp{78.42}{4.97}$ &
    $\musigp{89.15}{2.37}$ &
    $\musigp{67.54}{3.91}$ & $\musigp{98.60}{0.53}$ \\
    \hline

    $\mathcal{S} \rightarrow \mathcal{M}$ &
    $\pb{96.15}$  & $\musigp{85.21}{4.73}$ &
    $\musigp{84.63}{2.13}$ &
    $\musigp{85.19}{1.19}$ &
    $\musigpb{85.75}{1.20}$ &
    $\musigp{74.89}{2.60}$ &
    $\musigp{74.03}{1.50}$ &
    $\musigp{51.64}{2.77}$ & $\musigp{98.60}{0.53}$ \\
    \hline
\end{tabular}
}%\vspace{-3pt}%
\caption{Performance of our method and baselines in adapting models among \mnist ($\mathcal{M}$), \usps ($\mathcal{U}$), and \svhn ($\mathcal{S}$). 100 \images are trained for ten GD steps and three epochs. Our method outperforms few-shot domain adaptation~\citep{motiian2017few} and other baselines in most settings.
}
\vspace{-5pt}
\lbltbl{adapt}

%auto-ignore
%!TEX root = ../main.tex

\vspace{20pt}
\centering
\resizebox{0.88\columnwidth}{!}{%
\renewcommand{\arraystretch}{1.2}
\renewcommand{\psymbol}{}

\begin{tabular}{|c|c|c|c|c||c|}
    \hline
    \multirow{2}{*}{Target dataset} &
    \multirow{2}{*}{Ours} &
    \multirow{2}{*}{Random real} &
    \multirow{2}{*}{Optimized real} &
    \multirow{2}{*}{Average real} &
    \multirow{2}{*}{\shortstack{Fine-tune on \textbf{full}\\target dataset}} \\
    & & & & & \\
    \hline

    \pascalvocshort{2007} &
    $\pb{70.75}$ &
    $\musigp{19.41}{3.73}$ &
    $\musigp{23.82}{3.66}$ &
    $\p{9.94}$ &
    $\musigp{75.57}{0.18}$ \\
    \hline
    \birdsshort &
    $\pb{38.76}$ &
    $\musigp{7.11}{0.66}$ &
    $\musigp{7.23}{0.78}$ &
    $\p{2.88}$ &
    $\musigp{41.21}{0.51}$ \\
    \hline
\end{tabular}
}%\vspace{-3pt}
\caption{Performance of our method and baselines in adapting an \alexnet pre-trained on \imagenet to \pascalvoc{2007} and \birds. We use one \image per category, with one GD step repeated for three epochs. Our method significantly outperforms the baselines. We report results over $10$ runs.}
\lbltbl{adapt-imagenet}
% \vspace{-2pt}

\end{table}%

\myparagraph{\Fixed pre-trained weights on \imagenet.} In \reftbl{adapt-imagenet}, we adapt a widely used \alexnet model~\citep{krizhevsky2012imagenet} pre-trained on \imagenet \citep{deng2009imagenet} to  image classification on \pascalvoc{2007} \citep{everingham2010pascal} and \birds \citep{WahCUB_200_2011} datasets. Using only one \image per category, our method outperforms baselines significantly. Our method is on par with  fine-tuning on the full datasets with thousands of images.

\begin{figure}[t]
\centering
\input{figText/forget-numbers}
\end{figure}
\myparagraph{\Stoch pre-trained weights and a malicious data-poisoning objective.} \refsec{diff-obj} shows that our method can construct a new type of data poisoning, where an attacker can apply just one GD step with a few malicious data to manipulate a well-trained model. We train \images to make well-optimized neural networks to misclassify an attacked category as another target category \textit{within only one GD step}. Our method requires \textit{no} access to the exact weights of the model.
In \reffig{forget-attack}, we evaluate our method on $200$ held-out models, against various baselines using data derived from real images and incorrect labels. For baselines, we apply one GD step using the same numbers of images with modified labels (\ie, the attacked category images are labeled as target category) and report the highest overall accuracy \wrt the modified labels while misclassifying $\geq 10\%$ attacked category as target category. This avoids results with learning rates too low to change model behavior at all. While some baselines perform similarly well as our method on \mnist, our method significantly outperforms all the baselines on \cifar{10}.

%auto-ignore
%!TEX root = ../main.tex

\section{Discussion}
\lblsec{conclusion}
In this paper, we have presented \ours for compressing the knowledge of entire training data into a few synthetic training images. We can train a network to reach high performance with a small number of \images and several gradient descent steps. Finally, we demonstrate two extended settings including adapting pre-trained models to new datasets and performing a malicious data-poisoning attack.  In the future, we plan to extend our method to compressing large-scale visual datasets such as ImageNet and other types of data (\eg, audio and text). Also, our current method is sensitive to the distribution of initializations. We would like to investigate other initialization strategies, with which \ours can work well.

\myparagraph{Acknowledgments} This work was supported in part by NSF 1524817 on Advancing Visual Recognition with Feature Visualizations, NSF IIS-1633310, and Berkeley Deep Drive.

\newpage\clearpage

\small

\bibliography{reference}
\bibliographystyle{iclr2019_conference}

\newpage\clearpage
%auto-ignore
%!TEX root = ../main.tex

% \clearpage

\setcounter{section}{0}
\renewcommand{\thesection}{S-\arabic{section}}

\section{Supplementary Material}
\lblsec{supplement}

% \subsection{Experiment Details}
\lblsec{training_details}
In our experiments, we disable dropout layers in the networks due to the randomness and computational cost they introduce in distillation. Moreover, we initialize the distilled learning rates with a constant between $0.001$ and $0.02$ depending on the task, and use the Adam solver~\citep{kingma2014adam} with a learning rate of $0.001$. For \stoch initialization and \stoch pre-trained weights, we sample $4$ to $16$ initial weights in each optimization step. We run all the experiments on NVIDIA Titan Xp and V100 GPUs. We use one GPU for \fixed initial weights and four GPUs for \stoch initial weights. Each training typically takes $1$ to $4$ hours.

Below we describe the details of our baselines using real training images.
\begin{itemize}
\item \myparagraph{Random real images:} We randomly sample the same number of real images per category. We evaluate the performance over $10$ randomly sampled sets. %such set of sampled images are evaluated.
\item \myparagraph{Optimized real images:} We sample $50$ sets of real images using the procedure above, pick $10$ sets that achieve the best performance on $20$ held-out models and $1024$ randomly chosen training images, and evaluate the performance of these $10$ sets.
\item \myparagraph{$k$-means:} For each category, we use $k$-means clustering to extract the same number of cluster centroids as the number of \images in our method. We evaluate the method over $10$ runs. %such set of sampled images are evaluated.
\item \myparagraph{Average real images:} We compute the average image of all the images in each category, which is reused in different GD steps. We evaluate the model only once because average images are deterministic.
\end{itemize}

To enforce our optimized learning rate to be positive, we apply \texttt{softplus} to a scalar trained parameter.

\end{document}


\maketitle
\section{Implementation Details}

For all experiments, we initialize the networks using the method from \cite{he2015delving} since it allows better gradient flow regardless of network depth, which is useful in applying learned \images on different architectures. Furthermore, during training, we temporarily disable dropout and batch normalization layers due to two main reasons. First, randomness in the network makes learning distilled knowledge harder and slower. Second, backpropogating through the running statistics in batch normalization is computationally expensive.

\section{Experiment Details}

\begin{table}[h!]
\centering
\begin{tabular}{ |C{3cm}||C{2.2cm}|C{3.2cm}|  }
    \hline
    {\bf \lenet} & \multicolumn{2}{c|}{{\bf \largenet}} \\
    \hline
    $62006$ parameters & \multicolumn{2}{c|}{$204190$ parameters} \\
    \hline
    \hline
    {\tt Conv[C6, K5, P2]} & \multicolumn{2}{c|}{\tt Conv[C10, K5, P2]} \\
    \hline
    {\tt MaxPool[K2]} & \multicolumn{2}{c|}{\tt Conv[C10, K2, S2]} \\
    \hline
    {\tt Conv[C16, K5]} & \multicolumn{2}{c|}{\tt Conv[C20, K5]} \\
    \hline
    {\tt MaxPool[K2]} & \multirow{2}{2.2cm}{Residual connection \cite{he2016deep}} & {\tt Conv[C20, K5, P2]} \\
    \cline{1-1}\cline{3-3}
    {\tt FC[120, 84]} & & {\tt Conv[C20, K5, P2]} \\
    \hline
    {\tt FC[84, 10]} & \multicolumn{2}{c|}{\tt Conv[C30, K5, S2]} \\
    \hline
    & \multicolumn{2}{c|}{\tt FC[750, 200]} \\
    \hline
    & \multicolumn{2}{c|}{\tt FC[200, 120]} \\
    \hline
    & \multicolumn{2}{c|}{\tt FC[120, 10]} \\
    \hline
\end{tabular}
\vspace{5pt}
\caption{Architecture specifications for \lenet~\cite{lecun1998gradient} and \largenet. Each {\tt Conv} and {\tt FC} layer is followed by a {\tt ReLU} activation. For {\tt Conv} and {\tt MaxPool} layers, {\tt C} denotes number of output channels, {\tt K} kernel size, {\tt P} zero-padding size (default $0$), {\tt S} stride (default $1$). {\tt FC[X, Y]} means a fully-connected layer mapping {\tt X} features to {\tt Y} features. }
\lbltbl{mnist-arch-comp}
\end{table}

\section{Additional Experiment Results}
Figure 1-5 visualize the distilled images under various experiment settings. The captions are self-contained.
\begin{figure}
    \includegraphics[trim=40 0 0 0, clip, scale=0.28]{figures/charge_fixed_Cifar10.png}
    \caption{\Images from $4$ GD steps and corresponding trained learning rates, trained for a \vgg{11}~\cite{simonyan2013deep} with a particular known random initializations on \cifar{10}~\cite{cifar10}. This loads the particular initialization's test accuracy from $10\%$ to $33.48\%$.}
\end{figure}
\begin{figure}
    \includegraphics[trim=40 0 0 0, clip, scale=0.28]{figures/charge_random_rev_MNIST.png}
    \caption{\Images from $10$ GD steps and corresponding trained learning rates, trained using \lenet with unknown random initializations on \mnist. }
\end{figure}
\begin{figure}
    \includegraphics[trim=40 0 0 0, clip, scale=0.28]{figures/charge_random_rev_fixLR_MNIST.png}
    \caption{\Images from $10$ GD steps, trained with fixed learning rate $0.01$ using \lenet with unknown random initializations on \mnist. }
\end{figure}
\begin{figure}
    \includegraphics[trim=40 0 0 0, clip, scale=0.28]{figures/charge_random_rev_Cifar10.png}
    \caption{\Images from $10$ GD steps and corresponding trained learning rates, trained using \vgg{11} with unknown random initializations on \cifar{10}. }
\end{figure}
\begin{figure}
    \includegraphics[trim=40 0 0 0, clip, scale=0.28]{figures/charge_random_rev_fixLR_Cifar10.png}
    \caption{\Images from $10$ GD steps, trained with fixed learning rate $0.01$ using \vgg{11} with unknown random initializations on \cifar{10}. }
\end{figure}

\newpage\clearpage

\small

\bibliography{reference}
\bibliographystyle{plainnat}